\pdfoutput=1
\documentclass[11pt]{article}

\usepackage[final]{acl}
\usepackage{footmisc}
\usepackage{times}
\usepackage{latexsym}
\usepackage{subcaption}
\usepackage[T1]{fontenc}

\usepackage[utf8]{inputenc}

\usepackage{microtype}
\usepackage{booktabs}

\usepackage{inconsolata}
\usepackage{multirow}
\usepackage{array}
\usepackage{colortbl}     
\usepackage{xcolor}       

\usepackage{graphicx}
\usepackage{caption}
\usepackage{listings}
\usepackage{colortbl}
\usepackage{adjustbox}
\usepackage[linguistics]{forest}
\usepackage{todonotes}
\makeatletter
\newcommand*\iftodonotes{\if@todonotes@disabled\expandafter\@secondoftwo\else\expandafter\@firstoftwo\fi} 
\makeatother

\newcommand{\note}[4][]{\todo[author=#2,color=#3,size=\scriptsize,fancyline,caption={},#1]{#4}} 

\newcommand{\anna}[2][]{\note[#1]{anna}{green!40}{#2}}

\title{Do Syntactic Categories Help in Developmentally Motivated Curriculum Learning for Language Models?}

\author{
  Arzu Burcu Güven \quad Anna Rogers \quad Rob van der Goot \\
  IT University of Copenhagen, Denmark \\
  {\tt \{argy, arog, robv\}@itu.dk}
}

\begin{document}
\maketitle
\begin{abstract}




We examine the syntactic properties of BabyLM corpus, and age-groups within CHILDES. While we find that CHILDES does not exhibit strong syntactic differentiation by age, we show that the syntactic knowledge about the training data can be helpful in interpreting model performance on linguistic tasks. For curriculum learning, we explore developmental and several alternative cognitively inspired curriculum approaches. We find that some curricula help with reading tasks, but the main performance improvement come from using the subset of syntactically categorizable data, rather than the full noisy corpus. \footnote{https://github.com/arzuburcuguven/syntactic-categorization \label{repo-link}}

\end{abstract}

\section{Introduction}

Curriculum Learning (CL), a training regimen where the input is ordered from easier to more difficult, has been shown to improve performance of the machine learning algorithms in various scenarios \cite{soviany_curriculum_2022}.
In NLP, the BabyLM challenge \cite{warstadt_findings_2023}, inspired by human efficiency in acquiring language from a small amount of data, has sparked interest in applying CL to small-scale training setups. Most studies in this research area base their curricula on language or syntactic complexity. However, to quantify these complexities they rely on coarse proxies, such as ordering different corpora \cite{martinez_climb_2023}, mean length of utterance (MLU) \cite{oba_babylm_2023} or the average number of syntactic dependents \cite{mi_mmi01_2023}. Despite being a popular approach, CL has not consistently led to performance gains in these settings \cite{hu_findings_2024}.

One of the core corpora in CL studies in NLP is CHILDES \cite{macwhinney_childes_2000}, which consists mostly of interactions between children and adults. It is currently the primary resource for Child Directed Speech (CDS), which is known to exhibit distinct topical, lexical and morphosyntactic features \cite{gallaway_input_1999,  huttenlocher_language_2002, soderstrom_beyond_2007}. Several studies use CHILDES as a stand-in for developmentally grounded training \cite{feng_is_2024, huebner_order_2018, huebner_babyberta_2021, martinez_climb_2023}. Surprisingly, although there are many CL studies relying on CHILDES (based on CDS \cite{huebner_order_2018, huebner_babyberta_2021}, syntactic complexity \cite{oba_babylm_2023, mi_mmi01_2023}, or language complexity \cite{martinez_climb_2023}), its syntactic properties have not been explored in a fine-grained manner in this line of work.

\renewcommand{\arraystretch}{1}
\begin{table*}[ht]
\small
\centering
\begin{tabular}{p{0.15\textwidth} p{0.25\textwidth} p{0.50\textwidth}}
\toprule
\textbf{Macro-category} & \textbf{Syntactic Category} & \textbf{Examples} \\
\midrule
\multirow{7}{*}{\parbox[t]{\linewidth}{\centering \textbf{Simple}}}
& Subject-Verb & \textit{She runs. She opens the bottle.} \\
& Adverbs \& Possessives & \textit{Try again. She runs fast. She opens your bottle.} \\
& Prepositions & \textit{Good for you. She runs with her friend.} \\
& Particle verbs & \textit{Cut it off. She opens up to you.} \\
& Auxiliaries & \textit{She can run fast. She should open up to you.} \\
& Negation & \textit{Don’t run fast. She should not open up to you.} \\
& Tense & \textit{You are running fast. She has been opening up to you.} \\
\arrayrulecolor{gray!50}  
\specialrule{0.3pt}{0pt}{0pt}
\multirow{5}{*}{\parbox[t]{\linewidth}{\centering \textbf{Complex}}}
& Embedded clauses & \textit{Let’s go. I know what I need.} \\
& To-infinitives & \textit{I want to run. I’m going to call you.} \\
& Linked clauses & \textit{I want to run and smell flowers. I run because I like it.} \\
& Relative clauses & \textit{The tooth fairy who loves good children} \\
& Fragments & \textit{Uh, ah yes, umm, not into that} \\
\specialrule{0.3pt}{0pt}{0pt}
\multirow{1}{*}{\parbox[t]{\linewidth}{\centering \textbf{Interrogatives}}}
& Interrogatives & \textit{What? Is that a hat? Does she know what the moon is?} \\
\arrayrulecolor{black}  
\bottomrule
\end{tabular}
\caption{Developmental macro-categories, associated syntactic categories, and example utterances.}
\label{tab:tregex-groups}
\end{table*}

\begin{table}[t]
\centering
\small
\begin{tabular}{l@{\hskip 10pt}l@{\hskip 10pt}r}
\toprule
\textbf{Corpus} & \textbf{Genre} & \textbf{Tokens} \\
\midrule
CHILDES         & Child-directed speech & 25.9M \\
BNC Spoken           & Spoken English  & 9.2M \\
OpenSubtitles   & Movie subtitles & 25.8M \\
Switchboard     & Telephone conversations & 1.6M \\
Simple Wikipedia& Encyclopedia     & 17.3M \\
Gutenberg       & Children stories        & 31.0M \\
\bottomrule
\end{tabular}
\caption{Overview of corpora used in this study, with genre and token count after clean-up.}
\label{tab:corpus_stats}
\end{table}

To address the gaps in the literature, namely the lack of concrete curriculum quantification and the limited analysis of CHILDES both in itself and in comparison to other corpora as training data, we propose a syntax-based approach. Our contributions are as follows:

\begin{enumerate}
    \item We introduce a toolkit\footref{repo-link} to analyze, label, and order training data based on the syntactic properties of each sentence, based on approximately 300 expert-designed tregexes capturing 71\% of sentences in CHILDES. 
    \item We contribute a detailed analysis of the BabyLM corpora for syntactic properties, and we present the analysis of developmentally motivated marco-categories across each subcorpus. 
    \item For CHILDES, we examine distributions by age group. We find no clear differences that align with the developmental syntactic stages proposed in language acquisition research, and we propose hypotheses for why this might be the case.
    \item We train language models on syntactically and developmentally motivated curricula and compare them against baselines. We find that the primary performance gain stems not from CL itself, but from using syntactically categorizable data.
    \item We utilize our syntactic classification framework to compile syntactically isolated subcorpora, and conduct a study on cross-construction generalization. We observe mixed results: simpler categories do not cross-generalize, whereas more complex categories can improve performance on other complex ones.
    
\end{enumerate}



\section{Methods}

Our overall curriculum design is built upon classifying data by syntactic categories, and ordering the classified data according to curricula. We begin by describing the datasets used in this study, followed by the syntactic categories, the categorizing process, and the curriculum design.



\subsection{Datasets}

Both the training and the data analysis are conducted on the strict BabyLM dataset \cite{charpentier_babylm_2025}. The dataset comprises corpora with diverse properties, including CHILDES as CDS; Switchboard \cite{godfrey_switchboard_1992}, the spoken portion of the British National Corpus (BNC) \cite{20.500.14106/2554}, and OpenSubtitles \cite{lison_opensubtitles2016_2016} as adult-directed speech (ADS); and Simple English Wikipedia and Project Gutenberg (children stories) \cite{gerlach2018standardizedprojectgutenbergcorpus} as written text.

We remove speaker labels from all corpora, as the labels decrease the parser accuracy. For CHILDES, we additionally remove annotations and normalize nonstandard expressions. Sentence segmentation is applied to all corpora, and each resulting line is treated as a unit for parsing and extraction. We remove utterances shorter than two tokens. Table~\ref{tab:corpus_stats} summarizes the features and size of each corpus.

\begin{figure*}
    \begin{minipage}[t]{0.45\textwidth}
    \vspace*{0.6cm}
\begin{adjustbox}{max width=\columnwidth}
\begin{forest}
for tree={
  font=\footnotesize,         
  l sep=9pt, s sep=6pt,       
  inner sep=1pt,              
  align=center}
[S
  [NP [DT [My]] [NNS [feet]]]
  [VP
    [VBP [are]]
    [ADJP [JJ [dry]]]
    [SBAR
      [IN [because]]
      [S
        [NP [PRP [I]]]
        [VP [VBP [have]] [NP [NNS [boots]]]]
      ]
    ]
  ]
]
\end{forest}
\end{adjustbox}
\vspace*{0.6cm} 
\subcaption{Constituency parse of the sentence ``\textit{My feet are dry because I have boots}.''}
\label{fig:because-tree}
    \end{minipage}
    \hfill
    \begin{minipage}[t]{0.45\textwidth}
\begin{lstlisting}[language={}]
% Subject-verb or intransitive sentence:
(S
  [ <1 (NP <: /NN|DT|PRP|CD|FW|VBG|EX|WP/)
  | <1 (NP <1 /NN|DT|PRP|CD|UH|FW|VBG|WP/
          <2 /^NN|DT|PRP|CD|FW|WP/ !<3 __)
  | <1 (NP <1 /NN|DT|PRP|CD|UH|FW|WP/
          <2 /^NN|DT|PRP|CD|FW|VBG|WP/
          <3 /^NN|DT|PRP|CD|FW|WP/ !<4 __)
  ]
  <2 (VP <: /^VB/)
  <3 /^(\.|\.\.\.|!|\?)$/ 
)
!> __

% Wh-question (e.g.,  Who is talking to you?):
SBARQ<(/WH/$++(/SQ|S/<1(/VB|MD/)<2VP))

% Subordinating conjunction 
(e.g., My feet are dry because I have boots):
(NP!<<CC)
$++(VP<(/VB/
$++(SBAR<(/IN|WH/$++(S<NP<VP!<<CC)))))
\end{lstlisting}
\subcaption{Tregex Patterns needed to match the sentence ``\textit{My feet are dry because I have boots}.''}
\label{lst:tregex-patterns}
    \end{minipage}
\caption{Example of syntactic annotation (a) and tregexes (b) used to filter CHILDES}    
\end{figure*}

\subsection{Syntactic Categorization}

In order to design the syntactic categories, we examined various resources that classify syntactic phenomena into overarching groups, including typological databases such as Grambank \cite{lesage_overlooked_2022}, language universals \cite{croft_typology_2002}, and grammatical frameworks such as dependency relations \cite{de_marneffe_universal_2021}, and LinGO Grammar Matrix \cite{bender_grammar_2010}. Despite differences in terminology, underlying assumptions, and goals across the frameworks, we curated a set of categories that are at least represented twice among them. We found that the most comprehensive list was presented by Grambank, to which our 13 categories are most closely aligned. We restricted our final set to categories applicable to English. The resulting 13 categories are listed in Table~\ref{tab:tregex-groups}. For a further discussion of these categories, see Appendix~\ref{sec:appendix-categories}.

For parsing the corpora we used \citet{kitaev_constituency_2018}'s a constituency parser  for its ease of use and high performance. Data was analyzed using Tregex \cite{levy-andrew-2006-tregex}, for which we designed approximately 300 regular expressions targeting the sentences that can be categorized into the 13 syntactic categories. 
These expressions were crafted by an experienced syntactician with a graduate degree in computational linguistics and six years of professional experience in linguistics. Matches returned by the expressions are saved and reordered to curate corpus subsets. This setup also allows for corpus-specific or cross-corpus categorization. Extracted data can also be used to create filtered training data, for example, by excluding fragments or only including relative clauses.

Figure~\ref{fig:because-tree} shows a constituency tree of a complex sentence and Figure~\ref{lst:tregex-patterns} shows examples of Tregex patterns used to match the syntactic trees to different categories.

To the best of our knowledge, our Tregex patterns constitute the most extensive syntactic analysis of CHILDES to date; prior parsing studies used much smaller subsets (~65k-236k tokens; \citep{sagae_high-accuracy_2007, liu_data-driven_2023, yang_ud-english-childes_2025}). Even so, it categorizes only 71\% of sentences in the English portion of CHILDES, primarily because of the long tail of rare that would be impossible to fully cover with Tregexes and presence of noisy disfluencies (stutters, restarts, fillers), e.g., ``y you know b build this like real big thing to hold t planets from colliding together.''

\subsection{Curriculum}
\label{sec:Curriculum}
Most studies on language acquisition in English-speaking children focus on a specific syntactic phenomenon or developmental period. For instance, the seminal work by \citet{brown_first_2013} describes the acquisition of a variety of phenomena such as tense, possessives, and auxiliaries, yet omits others such as interrogatives and conjunctions. Similarly, \citet{braine_childrens_1976} focus exclusively on the first word combinations. Many studies approach acquisition from a universalist perspective, highlighting similarities among different language speakers \cite{slobin_crosslinguistic_1987}.\footnote{For numerous language-specific studies, see the series \textit{The Crosslinguistic Study of Language Acquisition} (ed. D.~I.~Slobin).}

However, to create a syntactically grounded developmental curriculum, we need a more comprehensive framework representing a wider range of phenomena. To this end, we adopted the developmental stages proposed by \citet{friedmann_stages_2021}, based on observations of 54 Hebrew-speaking children aged 1.5 to 6 years. These stages have also been applied to English to examine whether similar learning trajectories are also observed in the learning behavior of LMs \cite{evanson_language_2023}.

\citet{friedmann_stages_2021} identify three main stages in syntactic development: the first stage corresponds to simple subject–verb constructions, the second to interrogatives, and the third to relative clauses and embedded structures such as infinitives. We adopt these three stages as the basis for our main curriculum, labeling them as simple, interrogative, and complex. The 13 syntactic categories are mapped to these macro-categories as shown in Table~\ref{tab:tregex-groups}. 

We stress that this is only one possible hypothesis about how an effective curriculum could be constructed, and any conclusions would be made only with respect to it rather than developmentally motivated CL in general.

\subsection{Evaluation}

We evaluated our models using the shared BabyLM evaluation pipeline \cite{charpentier_babylm_2025}. Model evaluation was conducted on the full test set, with the exception of the Age of Acquisition (AoA) Evaluation Benchmark \cite{chang_word_2022}. The evaluation suite includes BLiMP \cite{warstadt-etal-2020-blimp-benchmark}, EWoK \cite{ivanova2024elements}, COMPS \cite{misra_comps_2023}, (Super)GLUE \cite{wang-etal-2018-glue}, Entity Tracking \cite{kim_entity_2023}, WUG\_ADJ \cite{hofmann_derivational_2024}, WUG\_PAST \cite{weissweiler-etal-2023-counting}, and Reading (self-paced and eye-tracking) \cite{de_varda_cloze_2024}.

BLiMP is a linguistic evaluation suite and BLiMP Supplement includes tasks specifically designed for BabyLM. COMPS and EWoK are world-knowledge datasets: COMPS focuses on immutable properties and their inheritance to subordinate concepts, whereas EWoK targets more dynamic, context-dependent properties. The Entity Tracking task assesses a model’s ability to follow the states of discourse entities. WUG\_ADJ evaluates adjective nominalization on nonce words, while WUG\_PAST assesses past-tense formation on nonce words. The Reading task measures the alignment between LM predictions and human processing through comparison with reading times. Lastly, GLUE is used for fine-tuning evaluation.

\subsection{A Closer Look into Datasets}
\label{sec:appendix-data}
This section provides an exposition of syntactic properties of corpora under study. First, we compare BabyLM sub-corpora and discuss differences in their distributions. Second, we examine age-ordered CHILDES to see whether syntactic distributions follow a developmental trajectory.

\subsubsection{Differences Among Corpora}
\begin{figure}[t]
  \centering
  \includegraphics[width=\linewidth]{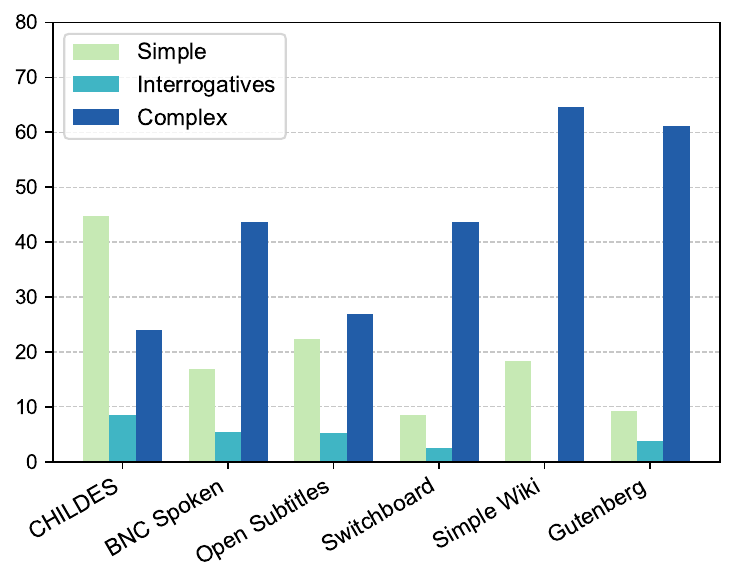}
  \caption{Distribution of macro-categories across corpora. Y-axis shows the percentage of sentences in each macro-category relative to the total number of sentences in the corpus.}
  \label{fig:syntax-stacked}
\end{figure}

In Figure~\ref{fig:syntax-stacked} we present the ratio of sentences that fall under each of the macro-categories for six different corpora. Here we can see the effect of corpus genre clearly, CHILDES, being the only example of CDS 
 differs markedly from other BabyLM corpora: Simple constructions and interrogatives account for 49\% of CHILDES, compared to 10.7–27.2\% in the other corpora. Among ADS corpora, BNC Spoken and Open Subtitles lean toward simpler language (16.1\% simple and 5.2\% interrogatives for the former; 22.0\% simple and 5.2\% interrogatives for the latter), whereas Switchboard has the lowest ratio of simple sentences (8.3\%) and a distribution more closely aligned with text corpora. 
 
 Among written corpora, Simple English Wikipedia has the lowest proportion of interrogatives (0.04\%), while Project Gutenberg is the most complex-leaning corpus, containing the highest proportion of complex sentences (59.8\%).

These distributions can be useful in interpretation of model performance as identifying which constructions are rare or overrepresented in the training data provides insight into model performance across different constructions.
For instance, \citet{huebner_babyberta_2021} suggest that the high frequency of questions in CHILDES may explain why models trained on it perform better on interrogatives. Indeed, among the corpora analyzed here, it has the highest proportion of interrogatives (7.8\%). \citet{padovani_child-directed_2025} compare models trained on CHILDES and Wikipedia. They evaluate the models on various agreement pairs and find that models trained on Wikipedia tend to perform better. This result is aligned with the distributions as relative clauses, which are one of the most challenging agreement distractors, are very scarce in CHILDES, amounting to only 0.8\% of the data whereas in Simple English Wikipedia, relative clauses account for the 11.5\% of the data, providing much richer training signal in terms of distractors.

\subsubsection{Age-Ordered CHILDES}

It is well-established that CDS is markedly different from ADS. One reason for this divergence is that adults adjust the syntactic complexity of their speech to match the child’s level of comprehension \cite{snow_mothers_1972, iii_childrens_1977}. Prior studies show that the syntactic complexity of CDS tends to increase over time, and that these changes in input correlate with children's language growth \cite{huttenlocher_sources_2010, silvey_effects_2021}. Given the relationship between CDS and the child's linguistic ability, we hypothesized that the age-ordered CHILDES would reflect the syntactic development of children.

Few studies have examined the differences between the age groups within CHILDES. Among them, the most relevant to our work is \citet{bunzeck_richness_2025}, which utilizes the morphological annotations within CHILDES with a regex-based parser, and assigns each sentence to one syntactic group among six: subject-verb constructions, interrogatives, imperatives, copular clauses, complex sentences and fragments. Their results show a subtle tendency toward interrogatives in the earlier age groups and subject-verb constructions in the older ones. 

We plot the macro-categories over age groups in Figure~\ref{fig:childes-syntax}, the full results on the fine-grained categories are reported in Appendix~\ref{sec:appendix-categories}, Figure~\ref{fig:CHILDES-ao-full}.
Our results do not reveal a clear developmental pattern across age groups. In line with \citet{bunzeck_richness_2025}'s results, there is a subtle tendency toward interrogatives in the earlier age groups, highest being 17.5\% with 3 to 4 age group. Subject-verb constructions, on the other hand, follow a non-linear trajectory, they peak at 48.9\% in between the ages of 1 to 2, then decrease and increase again between the ages of 5 and 6. Excluding the preverbal group, complex constructions start from 14.6\% in 1 to 2 ages and increase to 23.8\% at 5 to 6. In agreement with \citet{soderstrom_beyond_2007}'s findings, the preverbal segment of the corpus is syntactically distinct with a surprisingly high proportion of complex constructions (21\%).

\begin{figure}
  \centering
  \includegraphics[width=\linewidth]{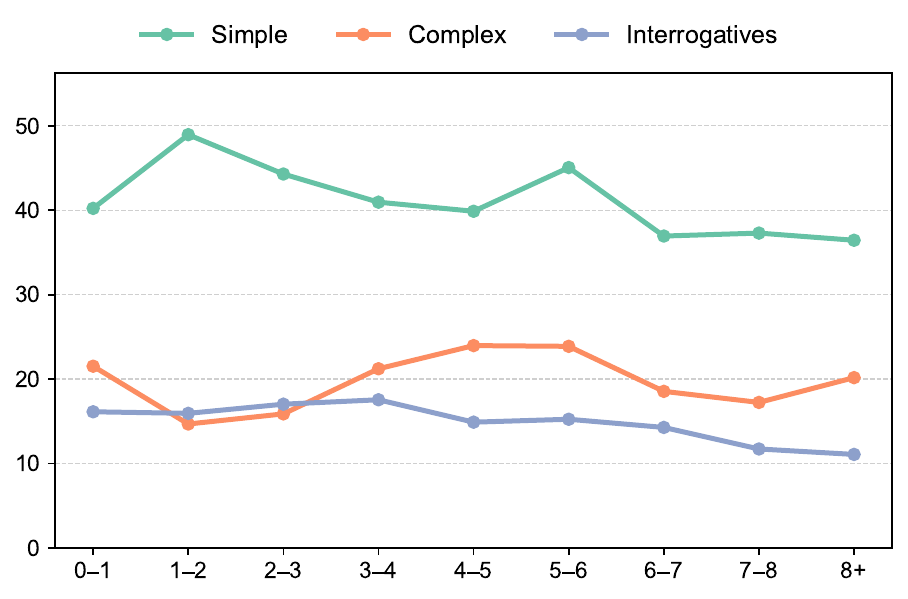}
  \caption{Distribution of macro-categories across age-ordered CHILDES. X-axis: age groups; Y-axis: percentage of sentences per macro-category.}
  \label{fig:childes-syntax}
\end{figure}

Our results suggest that CHILDES as a whole may not exhibit strong syntactic differentiation by age. Several factors likely contribute to this counter-intuitive outcome. The age groups aggregate data from 58 subcorpora, each containing transcripts from multiple children. Since children reach developmental milestones at individual rates \cite{bates_individual_2019}, it may be more informative to track syntactic development longitudinally for each child, as in \citet{brown_first_2013}. Socioeconomic status and dialect are also known to affect language complexity \cite{huttenlocher_language_2002}. Lastly, CHILDES transcripts come from different sessions, such as free play and book reading, which are known to differ in their syntactic characteristics \cite{bunzeck_richness_2025}.

\begin{table*}[t]
\centering
\setlength{\tabcolsep}{4pt}
\renewcommand{\arraystretch}{1.1}
\small
\begin{tabular}{cccccc}
\toprule
\textbf{Condition} & \textbf{BLIMP} & \textbf{SUPPLEMENT} & \textbf{EWOK} & \textbf{COMPS} & \textbf{GLUE} \\
\midrule
B1 & 70.24 $\pm$ 0.17 & \textbf{57.66 $\pm$ 0.10} & \textbf{50.53 $\pm$ 0.28} & \textbf{52.94 $\pm$ 0.49} & 57.12 $\pm$ 0.53 \\
B2 & \textbf{71.13 $\pm$ 0.62} & 52.98 $\pm$ 0.70 & 50.27 $\pm$ 0.18 & 51.74 $\pm$ 0.39 & 57.80 $\pm$ 0.74 \\
C1     & 69.88 $\pm$ 0.86 & 54.43 $\pm$ 2.04 & 50.08 $\pm$ 0.20 & 51.54 $\pm$ 0.58 & \textbf{57.83 $\pm$ 0.51} \\
C2     & 70.45 $\pm$ 0.72 & 55.85 $\pm$ 0.63 & 50.20 $\pm$ 0.18 & 51.19 $\pm$ 0.41 & 57.45 $\pm$ 0.41 \\
C3     & 70.98 $\pm$ 0.52 & 54.28 $\pm$ 0.29 & 50.06 $\pm$ 0.22 & 51.75 $\pm$ 0.80 & 57.62 $\pm$ 0.58 \\
C4     & 70.03 $\pm$ 0.60 & 53.09 $\pm$ 0.97 & 49.94 $\pm$ 0.26 & 51.31 $\pm$ 0.09 & 57.80 $\pm$ 0.35 \\
C5     & 70.44 $\pm$ 0.48 & 54.40 $\pm$ 0.89 & 50.19 $\pm$ 0.16 & 51.36 $\pm$ 0.43 & 57.61 $\pm$ 0.70 \\

\bottomrule
\end{tabular}
\caption{Mean $\pm$ SD (over seeds) for BLiMP, Supplement, EWOK, COMPS, and GLUE. Best per column in \textbf{bold}.}
\label{tab:results_blimp_supp_ewok_comps_glue}
\end{table*}

\begin{table*}[t]
\centering
\setlength{\tabcolsep}{4pt}
\renewcommand{\arraystretch}{1.1}
\small
\begin{tabular}{cccccc}
\toprule
\textbf{Condition} & \textbf{ENTITY} & \textbf{WUG\_ADJ} & \textbf{WUG\_PAST} & \textbf{READING\_SPR} & \textbf{READING\_ET} \\
\midrule
B1 & 20.70 $\pm$ 6.09 & 51.10 $\pm$ 7.76 & \textbf{2.28 $\pm$ 7.98} & 0.04 $\pm$ 0.05 & 0.42 $\pm$ 0.08 \\
B2 & \textbf{41.24 $\pm$ 1.21} & \textbf{68.87 $\pm$ 1.63} & -15.81 $\pm$ 6.08 & 0.14 $\pm$ 0.05 & 0.48 $\pm$ 0.17 \\
C1     & 32.34 $\pm$ 6.65 & 65.12 $\pm$ 1.67 & -19.89 $\pm$ 10.64 & \textbf{0.17 $\pm$ 0.05} & 0.64 $\pm$ 0.16 \\
C2     & 31.68 $\pm$ 8.75 & 62.06 $\pm$ 3.70 & -22.81 $\pm$ 5.26 & 0.15 $\pm$ 0.07 & \textbf{0.65 $\pm$ 0.12} \\
C3     & 38.76 $\pm$ 2.53 & 67.51 $\pm$ 1.10 & -15.71 $\pm$ 6.15 & 0.08 $\pm$ 0.04 & 0.42 $\pm$ 0.06 \\
C4     & 37.76 $\pm$ 3.71 & 66.84 $\pm$ 3.28 & -24.32 $\pm$ 3.86 & 0.05 $\pm$ 0.03 & 0.35 $\pm$ 0.08 \\
C5     & 37.83 $\pm$ 4.43 & 65.52 $\pm$ 4.60 & -22.73 $\pm$ 2.13 & 0.12 $\pm$ 0.07 & 0.39 $\pm$ 0.04 \\

\bottomrule
\end{tabular}
\caption{Mean $\pm$ SD (over seeds) for entity tracking, WUG, and reading metrics. WUG\_PAST column shows correlation results multiplied by 100. Best per column in \textbf{bold}.}
\label{tab:results_entity_wug_reading}
\end{table*}

\section{Experiments}

For both CL and generalization studies, we trained a model with the GPT-2 small architecture (124M parameters) \cite{radford2019language} from scratch using the Hugging Face Transformers library \cite{wolf-etal-2020-transformers}. Hyperparameters are detailed in Appendix~\ref{appendB}.

\subsection{Experiment 1: Curriculum}

\subsubsection{Methodology}

\begin{table}[t]
\centering
\footnotesize
\begin{tabular}{llp{0.55\linewidth}} 
\toprule
\textbf{Cond.} & \textbf{Tokens} & \textbf{Data order} \\
\midrule
B1 & 131M & Random \\
B2 & 77M & Random \\
C1 & 77M & S$\to$I$\to$C \\
C2 & 77M & S$\to$C \\
C3 & 77M & S$\to$C (gradual) \\
C4 & 77M & 80\% SIC, 20\% Mixed \\
C5 & 77M & 20\% Mixed, 80\% SIC, 20\% Mixed \\
\bottomrule
\end{tabular}
\caption{
Summary of training conditions. S=Simple, I=Interrogatives, C=Complex.
}
\label{tab:curricula}
\end{table}

This section describes experiments in which training sets are organized according to different curriculum approaches. The research question we address is "Does training on a developmentally motivated syntactic curriculum improve LM performance compared to random ordering or other curriculum variants?" To this end we train seven models: two baselines (B1, B2) and five curriculum variants (C1–C5). Table~\ref{tab:curricula} summarizes all training conditions.

The baselines are B1, the full BabyLM corpus in random order, and B2, an extracted subset of BabyLM corpus containing the union of all syntactically categorized data in random order. C1 (developmental curriculum, Section~\ref{sec:Curriculum}) groups the syntactically categorized training data into simple, interrogative, and complex stages, shuffling within each stage before concatenating them to form the final corpus.

To contrast with the developmentally grounded approach, we also devise several alternative curricula. In the simple-to-complex curriculum (C2), we categorize each syntactic structure as either simple or complex based on the presence of nested embedding. We then concatenate these two subgroups. In C3, we use the same simple and complex division described above but interleave them such that the dataset starts from only simple examples, progresses to a balanced dataset and ends with only complex examples. To achieve this, we employ a probabilistic sampling function that decreases the probability of sampling from the simple dataset and increase the probability of sampling from the complex dataset over the course of the sampling process. 

The last two CL approaches are inspired by the Learn–Focus–Review (LFR) strategy of \citet{prakriya-etal-2025-accelerating}, a cognitively inspired dynamic learning paradigm. In the initial learn phase, models see a portion of randomly sampled training data. In the focus phase, more challenging portions of the data are clustered, and in the review phase, the remaining data is reintroduced to prevent forgetting. For C4, 20\% of the syntactically labeled data is held out, the remaining 80\% is constructed as in C1, and the held-out portion is appended as a review at the end. For C5, 40\% of the data is held out, 60\% is constructed as in C1, and the held-out portion is split in half, with one half appended to the beginning and the other half to the end of the corpus.

\begin{table*}[t]
\centering
\setlength{\tabcolsep}{5pt}
\renewcommand{\arraystretch}{1.0}
\begin{tabular}{cccccc}
\toprule
\textbf{Condition} &
\textbf{Hypernym} &
\textbf{QA\_easy} &
\textbf{QA\_tricky} &
\textbf{SubjAuxInv} &
\textbf{Turn\_taking} \\
\midrule
B1 & 48.99 $\pm$ 0.35 & \textbf{55.47 $\pm$ 2.71} & \textbf{39.55 $\pm$ 1.04} & 84.02 $\pm$ 1.32 & \textbf{60.27 $\pm$ 2.17} \\
B2 & 49.82 $\pm$ 0.50 & 49.61 $\pm$ 2.66 & 27.88 $\pm$ 3.39 & 87.68 $\pm$ 1.81 & 49.91 $\pm$ 1.94 \\
C1 & 50.27 $\pm$ 0.68 & 52.34 $\pm$ 4.51 & 36.21 $\pm$ 2.29 & 83.77 $\pm$ 5.53 & 49.55 $\pm$ 0.45 \\
C2 & 49.94 $\pm$ 1.08 & 52.73 $\pm$ 3.46 & 38.03 $\pm$ 3.14 & \textbf{87.95 $\pm$ 0.66} & 50.62 $\pm$ 1.28 \\
C3 & 50.23 $\pm$ 1.52 & 53.90 $\pm$ 1.57 & 29.39 $\pm$ 1.61 & 87.20 $\pm$ 1.80 & 50.62 $\pm$ 0.68 \\
C4 & 49.47 $\pm$ 1.10 & 50.00 $\pm$ 2.21 & 30.00 $\pm$ 1.60 & 85.88 $\pm$ 0.70 & 50.09 $\pm$ 1.18 \\
C5 & \textbf{50.62 $\pm$ 0.91} & 52.73 $\pm$ 1.97 & 31.06 $\pm$ 1.94 & 88.54 $\pm$ 1.07 & 49.02 $\pm$ 1.38 \\
\bottomrule
\end{tabular}
\caption{Mean $\pm$ SD over seeds for UID subtasks. Best per column in bold.}
\label{tab:uid_breakdown}
\end{table*}

\subsubsection{Results}

We report averaged results over four seeds on the BabyLM test suite in Table~\ref{tab:results_entity_wug_reading} and Table~\ref{tab:results_blimp_supp_ewok_comps_glue}.
While curriculum learning offers some task-specific benefits, the main finding is that models trained on parsed and categorized data perform on par with the B1 baseline despite requiring 40\% fewer training steps. B1 still leads on BLiMP Supplement, EWOK, COMPS and WUG\_PAST, though the EWOK and COMPS margins are small. 

The difference in BLiMP Supplement scores may stem from a preprocessing decision: to make our training data parser-compatible, we removed speaker labels. As a result only the B1 model, which was trained on the whole BabyLM corpus, was shown examples with speaker labels.
As shown in Table~\ref{tab:uid_breakdown} (Appendix~\ref{appendB}), B1’s higher Supplement score is concentrated in three subcategories, \textit{QA\_easy}, \textit{QA\_tricky}, and \textit{turn-taking}; each containing speaker labels.
Since in the main BLiMP benchmark and other Supplement categories the other models outperform B1, this suggests that presence of the speaker labels likely accounts for the observed gap.

The difference in performance on WUG\_PAST is more difficult to interpret. A qualitative analysis of the predictions shows that B1 models tend to apply regular inflection to wug words more often, aligning more closely with human data. In contrast, the other models more frequently produce irregular inflections, correlating negatively with the baseline. For the WUG\_ADJ task, however, B1 underperforms compared to all other models. One possible explanation is that cleaner data makes models more attentive to irregularities. This may be an advantage in tasks with a constrained prediction space, such as selecting from a limited set of adjective nominalizers, but a disadvantage in open-set tasks like WUG\_PAST.

For GLUE, entity tracking, and reading tasks, models trained on categorized data outperform the B1 models. Especially for reading tasks, both self-paced reading and eye-tracking, curriculum models C1 and C2 show the highest performance, suggesting that the curriculum approaches can provide a signal that shortens the gap between human and machine processing.

\subsection{Experiment 2: Generalization}

\subsubsection{Methodology}

\begin{table}[t]
\centering
\small
\begin{tabular}{p{0.28\linewidth} p{0.6\linewidth}}
\toprule
\textbf{Category} & \textbf{Constructions} \\
\midrule
Subject-Verb     & Subject-Verb patterns \\
Modifier          & Adverbs, Possessives, and Prepositions \\
Verbal            & Particle verbs, Auxiliaries, Negation, and Tense \\
Embedded C.  & Small clauses, reported speech\\
Infinitives       & Infinitives \\
Linked Clauses    & Coordination, Subordination \\
Relative Clauses  & Relative Clauses \\
Interrogatives    & Yes/no, wh-questions \\
\bottomrule
\end{tabular}
\caption{Syntactic category groups used in the generalization study and their corresponding constructions.}
\label{tab:generalization-groups}
\end{table}

In this experiment, each model is trained on a single category and evaluated on eight validation sets corresponding to the distinct eight categories given in Table~\ref{tab:generalization-groups}. We approach this task as a generalization study, using the perplexity values as a proxy for models' ability to learn both the category they trained on and the remaining seven unseen categories.

For each group, we sample 2M tokens for training and 200K tokens for validation from the syntactically classified portion of BabyLM corpus. Sampling is restricted to sentences matching the target group's criteria. We train GPT-2 small models from scratch on each subset for one epoch. All results are averaged over five random seeds.
\\

\begin{figure}[!t]
  \centering
  \includegraphics[width=\linewidth]{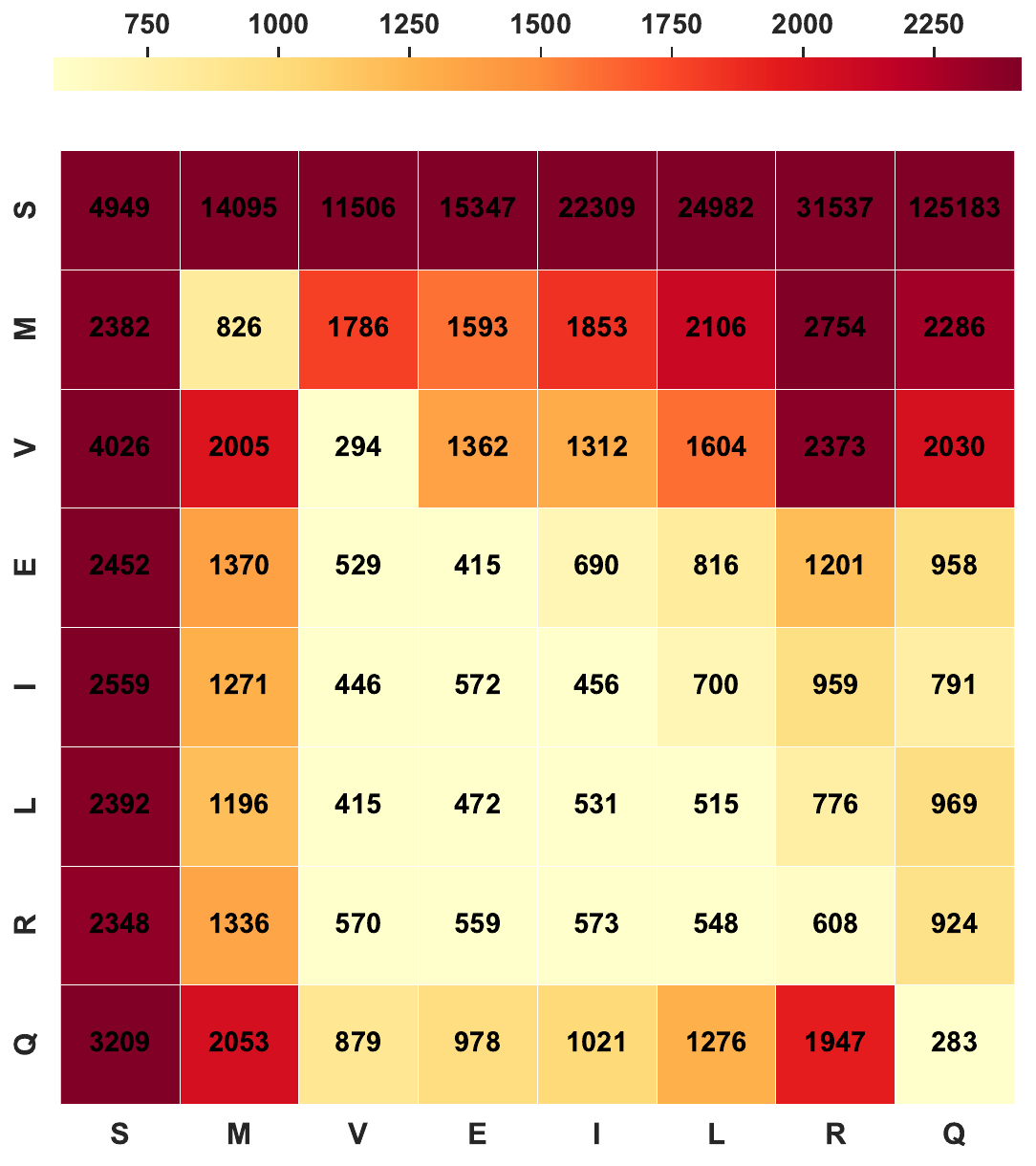}
  \caption{Cross-subset validation perplexity heatmap. Rows = training subset; columns = evaluation subset. Abbreviations: S=SVX, M=Modifiers, V=Verbal, E=Embedded, I=Infinitives, L=Coordination, R=Relative, Q=Question. Cell values are validation perplexities (lower is better).}
  \label{fig:generalization}
\end{figure}

\subsubsection{Results}

In Figure~\ref{fig:generalization} we report mean perplexity values across seeds. As expected, each model achieves its lowest perplexity when evaluated on the same syntactic category it was trained on (diagonal entries). Off-diagonal values indicate cross-category generalization.

Performance patterns vary across categories. The \textit{Subject–Verb} group (SVX) shows the largest drop in both in-category and cross-category performance, likely due to the high frequency of single-word (e.g., ``Run!'') and fragmentary utterances (e.g., ``all gone''). The \textit{Verbal} and \textit{Modifier} groups also generalize poorly.  Models trained on \textit{questions}, despite the data exhibiting unique syntactic patterns such as subject auxiliary inversion, generalize better than those trained on \textit{Subject–Verb}, \textit{Verbal} and \textit{Modifier} constructions. Models trained on complex constructions tend to generalize better to other complex categories. The \textit{Coordination}-trained model exhibits the strongest overall generalization, with the lowest mean off-diagonal perplexity (962.20) and the lowest perplexity on the mixed test set (574.2).

Overall perplexities remain high, and there is limited evidence for genuine syntactic generalization, particularly from simpler to more complex categories. Prior work demonstrating such transfer with transformer architectures typically relies on synthetic datasets with tightly controlled syntax and vocabulary \cite{murty_grokking_2023, ahuja_learning_2025, someya-etal-2024-targeted}. Our subsets are selected by syntactic criteria but retain naturalistic variation in sentence form and vocabulary. These results highlight the difficulty of isolating syntactic generalization in naturalistic data and suggest that stricter control of lexical and structural properties may be necessary for clearer conclusions.

\section{Conclusion}

This study contributes the most detailed syntactic analysis of BabyLM data to date, implemented as an open-source toolkit for analysing, labeling and ordering training data.\footref{repo-link} This enabled both modeling experiments and a systematic analysis of syntactic patterns in CHILDES, where, counter-intuitively, we find no clear differences in distributions that would align with syntactic stages proposed in language acquisition research. Likewise, we find that developmentally motivated curriculum has a modest effect in language model training, compared to simply training the models on a subset of training data filtered to only syntactically categorizable sentences. 

Efficient curriculum learning for language models that is inspired by human learning stages remains an elusive goal. The results of this study suggest that continued focus solely on syntax may be counter-productive, and that the noise in popular resources such as CHILDES may by itself have an outsized effect in studies relying on it.


\section*{Limitations}

We note the following limitations of this study: 

\begin{enumerate}
\item We did not observe developmental patterns in the aggregated CHILDES data, but our analysis did not extend to a more fine-grained level where confounding factors could be mitigated.
\item Our syntactic categorization covered 71\% of the BabyLM; some of the remaining gap is attributable to our data cleaning practices, but a portion remains unexplained. 
\item The absence of clear effects from CL or generalization may stem from several factors, and this study does not establish which ones are the most relevant. It is possible that isolating syntactic properties alone could be insufficient, or our method of isolation may not capture the most relevant distinctions. Alternatively, the targeted developmental progression and generalization may not be reproducible with the transformer architecture or training conditions used.
\end{enumerate}

\bibliography{Baby}


\appendix

\section{Category Details}
\label{sec:appendix-categories}

Below, we list our categories ordered by an increasing number of terminals and combinatorial possibilities. We start from simple noun phrases (NP), verb phrases (VP), adjective phrases (ADJP) and Subject-Verb constructions that can be built with them. For the categories with simpler constructions without any nested structures, the Tregex patterns match entire sequences and tightly constrain the contents of each node to exclude any complex expansions within the tree. For the more complex categories, we switch to partial matching, without constraining the preterminal nodes. 

\begin{itemize}

\item \textit{Subject-Verb Constructions}: For the sake of readability we use the term Subject-Verb Constructions, but the structures included are intransitive sentences (SV), transitive sentences (SVO), imperatives and copular sentences (SVC). Preterminals included in this category are simple NPs, VPs and ADJPs that have limited amount of nodes and no nested structures under them. Along with the well formed structures, we include sequences that consist of phrases such as \textit{Beautiful girl, the doll, all toys, love you Baby} etc. For the following categories up to the interrogatives, the sentence structures are limited to the ones described here.

\item \textit{Possessives and Adverbials}: For this category, we add POS and ADV preterminals to the former group. The NPs are extended to include possessives e.g., \textit{The girl's hat is beautiful}. Adverbial phrases are allowed both under VPs and directly under the S node.
    
\item \textit{Prepositions}: Phrases headed by PPs (\textit{at the table}), NPs governing over PPs (\textit{the girl with the blue ribbon}), ADJPs governing over PPs (\textit{good for you}) and VPs governing over PPs (\textit{walk to me}) are included both as standalone phrases and as participants in the SVX structures.

\item \textit{Particles}: VP categories are extended to include particle verbs (\textit{take off, put on}). This category forms one of the smallest categories in terms of how many sentences it captures, along with auxiliaries and tense.

\item \textit{Auxiliaries}: Here we repeat all the canonical sentence types from the former categories, SVX, SVX with adverbs, SVX with PPs and so on and modify the VPs to govern over an auxiliary. 

\item \textit{Negation}: The scope is again limited to all the canonical sentence types from the former categories and VPs are modified to govern over the negation particle.

\item \textit{Tense}: Although we have not differentiated between simple present or simple past tenses in the former categories, the more complex tenses such as progressive and perfective require a specific VP category. Again, we repeat all the canonical sentence types from the former categories, and modify the VPs to allow for the capture of complex tenses.

\item \textit{Interrogatives}: Here we include different types of interrogatives: Yes/no questions (\textit{Is she coming?}), Wh-questions (\textit{What is she doing?}), tag questions (\textit{She doesn't know, does she?}) and question fragments (\textit{What?, Did she?}).

\item \textit{Embedded Clauses}: This group captures a variety of nested structures in which at least two predicates are present. This includes let-constructions such as \textit{let me go}, causatives (\textit{I will make him bite mommy}) and small clauses (\textit{I think you can fix it}).

\item \textit{Infinitives}: This category captures the to-infinitives and gerunds e.g., \textit{She wants to drink from her cup.}

\item \textit{Clause Linking}: Here we include coordinating conjunctions (\textit{She ate an apple but the apple was rotten}) and subordinating conjunctions (\textit{My feet are dry because I have boots.}).

\item \textit{Relative Clauses}: This category is adapted from \citet{hsiao_nature_2023}, which includes relative clauses of subject (\textit{The man who kicked the ball}), object (\textit{the fun I had}) and passive (\textit{the houses that were built}) types.

\item \textit{Fragments}: While we allow phrase level constructions when they represent a well formed phrase, malformed phrases and interjections fall into this group.

\end{itemize}


\begin{figure*}[!t]
  \centering
  \includegraphics[width=\textwidth]{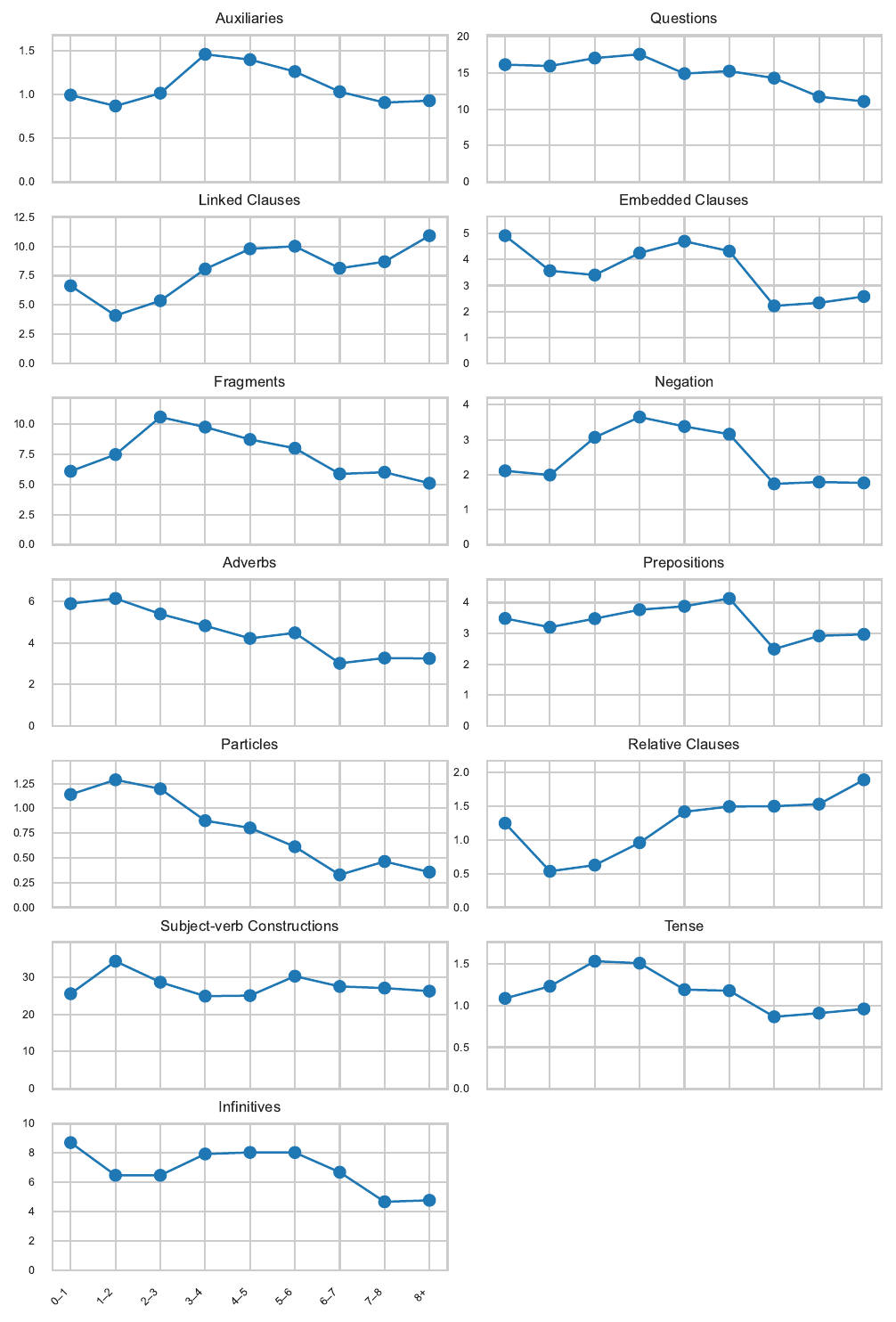}
  \caption{Percentage distribution of syntactic categories across age groups in CHILDES.}
  \label{fig:CHILDES-ao-full}
\end{figure*}

\section{Model Details}
\label{appendB}

We tuned hyperparameters with a sweep: learning rate sampled log-uniformly in
$[5\times10^{-6},\,5\times10^{-4}]$ and per-device train batch size $\in\{8,16,32\}$;
the best model was selected by validation-set perplexity. Remaining hyperparameters were taken from \citet{radford2019language}. The full set of hyperparameters is shown in \autoref{tab:hyperparams}.

\begin{table}[t]
\centering
\small
\begin{tabular}{ll}
\hline
\textbf{Hyperparameter} & \textbf{Value} \\
\hline
Model type & GPT-2 small \\
Parameters & 124M \\
Vocabulary size & 50,257 \\
Context size & 1024 \\
Dropout & 0.1 \\
Learning rate & 1.88$\times 10^{-4}$ \\
Scheduler & Linear \\
Weight decay & 0 \\
Epochs & 1 \\
Batch size & 8 \\
Optimizer & AdamW \\
\hline
\end{tabular}
\caption{Training hyperparameters for GPT-2 small}
\label{tab:hyperparams}
\end{table}


\end{document}